\documentclass[10pt,final,a4paper]{pj}

\usepackage{graphicx}
\usepackage[cmex10]{amsmath}
\usepackage{amssymb,bm}
\usepackage{algpseudocode} 
\newcommand{\algrule}[1][.5pt]{\par\vskip.25\baselineskip\hrule height #1\par\vskip.25\baselineskip}
\usepackage[caption=false,font=footnotesize]{subfig}
\usepackage{hyperref}

\newcommand{\ve}[1]{\mathbf{{#1}}}
\newcommand{\abs}[1]{\ensuremath{\vert #1\vert}}
\newcommand{\norm}[1]{\ensuremath{\Vert #1\Vert}}

\begin{document}
\setcounter{page}{1}
\pjheader{Vol.\ x, y--z, 2013}

%
\title[Scan-based Compressed Terahertz Imaging]{Scan-based Compressed Terahertz Imaging and Real-Time Reconstruction via the Complex-valued Fast Block Sparse Bayesian Learning Algorithm}

\author{Benyuan~Liu,~Hongqi~Fan,~Zaiqi~Lu and Qiang~Fu}

\address{The Science and Technology on Automatic Target Recognition Laboratory, National University of Defense Technology. Changsha, Hunan, P. R. China, 410074. E-mail: liubenyuan@gmail.com}%

\runningauthor{Benyuan}
\tocauthor{L.~B}

\begin{abstract}
	Compressed Sensing based Terahertz imaging (CS-THz) is a computational imaging technique. It uses only one THz receiver to accumulate the random modulated image measurements where the original THz image is reconstruct from these measurements using compressed sensing solvers. The advantage of the CS-THz is its reduced acquisition time compared with the raster scan mode. However, when it applied to large-scale two-dimensional (2D) imaging, the increased dimension resulted in both high computational complexity and excessive memory usage. 
In this paper, we introduced a novel CS-based THz imaging system that progressively compressed the THz image column by column. Therefore, the CS-THz system could be simplified with a much smaller sized modulator and reduced dimension. In order to utilize the block structure and the correlation of adjacent columns of the THz image, a complex-valued block sparse Bayesian learning algorithm was proposed. 
We conducted systematic evaluation of state-of-the-art CS algorithms under the scan based CS-THz architecture. The compression ratios and the choices of the sensing matrices were analyzed in detail using both synthetic and real-life THz images. Simulation results showed that both the scan based architecture and the proposed recovery algorithm were superior and efficient for large scale CS-THz applications. 
\end{abstract}


%
%
%
%
\section{Introduction}

\begin{figure}[!htb]
\centerline{\includegraphics[width=.5\linewidth]{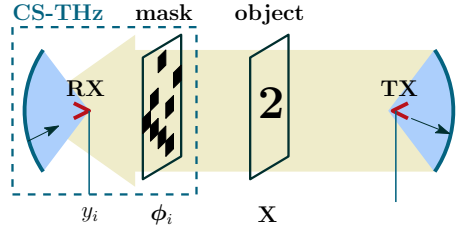}}
\caption{The illustration of the Compressed Terahertz Imaging System.%
The CS-THz system consists one receivers and a switching modulator placed in-between the receiver and the object.}
\label{fig:basic_cs_thz}
\end{figure}

The development of Compressed Sensing based Terahertz imaging\cite{Shen2012} follows the path of the single pixel camera\cite{Duarte2008}. The architecture of the CS-THz system is given in Fig. \ref{fig:basic_cs_thz}.
A typical CS-THz system consists of a single Terahertz Receiver (RX) and a switching modulator. The modulator is also called a \emph{mask} device. The object is illuminated by a Terahertz emitter (TX) and the propagation wave passes through a randomized masking device placed between the Terahertz receiver (RX) and the object. 
The masks are usually implemented using a metal plate\cite{Shen2012}, where at a given spatio location a hole that allows the transmission is denoted by $1$ and a block is $0$. At each snapshot, a mask is selected from a set of pre-built patterns. The propagation wave of the object is modulated and accumulated at the RX and a measurement is generated with such mask pattern. 

A THz image could be viewed as a complex-valued matrix, i.e., $\ve{X}\in\mathbb{C}^{N\times N}$, where $N$ denotes its spatio dimension in units of the beam-width of the focused beam. 
Traditionally with a single receiver, obtaining a full resolution THz image requires $N^2$ raster scans, which is time-consuming. CS-based Terahertz imaging can reduce the acquisition time by using only a few randomly modulated THz image measurements. The number of the measurements is denoted by $S$ and we usually have $S\ll N$. The original THz image is reconstruct from these measurements via Compressed Sensing (CS) solvers\cite{Berg2008}. 

\subsection{Problem Formulation}
We denote by $\bm{\phi}_i$ the modulated vector represents the pattern of a mask and $\ve{x}\triangleq\mathrm{vec}(\ve{X})$ the vectorized form of the THz image $\ve{X}$. The measurements $\ve{y}=\{y_1,\ldots,y_S\}$ are progressively obtained via a random projection of $\ve{x}$, i.e., $y_i = \langle\bm{\phi}_i,\ve{x}\rangle, \forall i$. Therefore, 
\begin{equation}\label{eq:cs_thz_old}
\ve{y} = \bm{\Phi}\ve{x}.
\end{equation}
where $\bm{\Phi}$ is a ${S\times N^2}$ sized sensing matrix with the $i$th row given by $\bm{\phi}_i$. 
The compression ratio (CR) is defined as,
\begin{equation}\label{eq:cs_cr_old}
	\mathrm{CR} = \frac{N^2 - S}{N^2}.
\end{equation}
The acquisition time of a full resolution THz image can be reduced by a factor of CR. The requirement number of the measurements is proportional to the sparsity of the THz image. When the image is sparse or can be sparsely represented in transformed domains, we could use much smaller measurements to recover the original image with high fidelity\cite{Candes2006,Candes2008a,li2013compressive}. 

However, a major drawback of the CS-THz imaging system is the {\it dimensional curse}. When it applied to large-scale two-dimensional (2D) imaging, the CS recovery algorithm needs to search the whole $N^2$ signal space to find the sparse solution for $\ve{X}$. For a typical application such as biomedical THz imaging\cite{tonouchi2007cutting,ohrstrom2010technical}, the dimension may exceed $100$, i.e., $N>100$. This results unrealistic storage and computational load to recover the original image real-time. The increased dimension also poses challenges to the design of the sensing matrix $\bm{\Phi}$. As each row of the sensing matrix corresponds to a masking device and is fabricated using either a steel plate\cite{Shen2012} or a metamaterial\cite{hunt2013metamaterial,kuznetsov2012matrix} with size $N^2$. Therefore, a small sized sensing matrix is preferred for practical applications.

The switching of the masks was usually done in a mechanical way, e.g, the rotated plate\cite{Shen2012}. In practical applications, a fast and electrical based THz imaging system is preferred. Recently, metamaterial\cite{hunt2013metamaterial,Reinhard2013} has been used as a novel masking device. The metamaterial device switches the masks electronically and it is equivalent to a complex-valued modulation vector. The CS-THz system can be built upon this new material. However, the analysis of the CS-THz system utilizing this new forms of masking devices is lacking. A systematic evaluation of the complex-valued sensing system using state-of-the-art CS recovery algorithms is needed to design a successful CS-THz system.

\subsection{Summary of Contributions}
In this paper we presented a novel scan-based CS-THz imaging system. In this system, we used a much smaller sized sensing matrix $\bm{\Phi}$ to compress the Terahertz image progressively in only one dimension. Each mask can be viewed as one dimensional (1D) random pattern of size $N$. The small-sized sensing matrix corresponds to smaller set of masks, which is easier to fabricate using either the metal plate or the metamaterial.

We observed that each column of the THz image usually exhibits a block sparse structure and the adjacent columns of the THz image are usually highly correlated. In order to exploit these underlying information, we provided an advanced CS recovery algorithm. The proposed algorithm was motivated by the Spatio-Temporal Block Sparse Bayesian Learning framework\cite{Zhang2013,Zhang2012a}. We extended the algorithm to handle the complex-valued signals and incorporate the Block Multiple-Measurement Vector (BMMV) model. We also provided a fast marginalized implementation to boost the speed.

We conducted extensive evaluations of the scan-based CS-THz architecture using both synthetic and real-life THz images. State-of-the-art CS algorithms were compared against the proposed algorithm under different compression ratios and different forms of the sensing matrices. In these experiments, we valid the use of scan-based CS-THz systems. We also showed that the proposed algorithm was the most efficient one and had improved recovery performance over traditional CS algorithms.

\subsection{Outline and Notations}
The rest of the paper is organized as follows. Section 2 gives an overview of the scan based CS-THz framework. Section 3 derives the complex-valued, fast BMMV modeled sparse Bayesian learning algorithm. In section 4, we conduct comprehensive experiments using both synthetic and real-life THz images. Conclusion is drawn in the last section.

Throughout the paper, {\bf bold} letters are reserved for vectors $\ve{x}$ and matrices $\ve{X}$. 
$\ve{x}_i$ and $\ve{x}^j$ denote the $i$th row and $j$th column vector of $\ve{X}$. $\ve{1}$ is a column vector of $1$s and $\ve{I}_N$ is an identity matrix with size $N$. 
\section{Scan-based Compressive Terahertz Imaging System}

\subsection{The Architecture of the Scan-based CS-THz System}
\begin{figure}[!ht]
\centerline{\includegraphics[width=.8\linewidth]{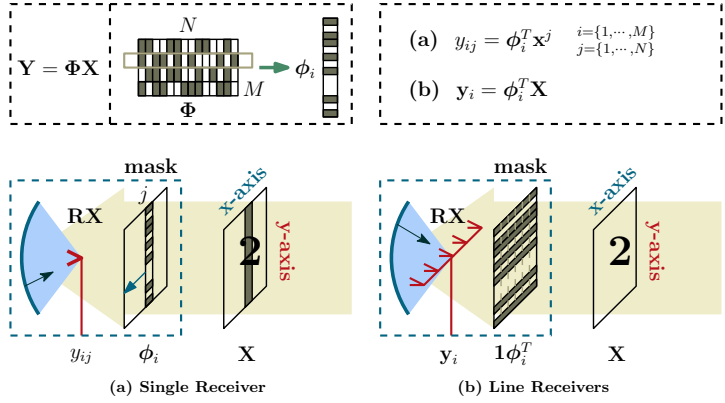}}
\caption{The scan-based CS-THz Imaging System. The system could be implemented using either (a) a single receiver and (b) a line receivers. When using a single receiver, the mask plate $\bm{\phi}_i$ moves along x-axis and iteratively senses the columns of the object $\ve{X}$. We could also use line receivers with an equivalent aperture mask pattern $\mathbf{1}\bm{\phi}_i^T$ to obtain the $i$th row of the compressed measurements simultaneously.}
\label{fig:arch_thz}
\end{figure}

The architecture of the scan-based CS-THz system is intuitive: we can sense the THz image progressively in only {\it one dimension}.
It could be implemented using either one single receiver or a receiver array with $N$ receivers lined up, as shown in Fig. \ref{fig:arch_thz}.

Let $\ve{x}^j$ denotes the $j$th column pixels of the image $\ve{X}$. Under the scan based CS-THz paradigm, it is compressed by multiplying a underdetermined matrix $\bm{\Phi}$ with size $M\times N$. Each row of $\bm{\Phi}$, denoted by $\bm{\phi}_i$, corresponds to a modulated vector, as shown in Fig. \ref{fig:arch_thz} (a). The $i$th modulated vector moved along the x-axis of the object $\ve{X}$ while at the $j$th scan, a measurement $y_i^j$ was generated via $y_i^j = \bm{\phi}_i\ve{x}^j$. The compressed measurements $\ve{Y}=\{y_i^j\}$ are obtained by iterative all mask vectors on each columns of the THz image,
\begin{equation}
	\ve{Y} = \{y_i^j\}, \quad i=\{1,\cdots,M\},~~j=\{1,\cdots,N\}.
\end{equation}
When using line receivers, as shown in Fig. \ref{fig:arch_thz} (b), the system could simultaneously acquire a row of compressed measurements,
\begin{equation}
	\ve{y}_i = \bm{\phi}_i \ve{X},
\end{equation}
and the acquisition time could be further reduced by a factor of $N$. 

\subsection{Related to the Traditional CS-THz System}
The scan based CS-THz system is mathematically equivalent to the traditional CS-THz system using the Kronecker sensing scheme,
\begin{equation}
	\bm{\Phi}'=\ve{I}_N\otimes\bm{\Phi},
\end{equation}
where $\bm{\Phi}'$ is the sensing matrix used in \eqref{eq:cs_thz_old} and $\otimes$ is the Kronecker operator.
The compression ratio of the proposed architecture is 
\begin{equation}
CR = \frac{N - M}{N}.
\end{equation}
The compression ratio of the traditional CS-THz was given in \eqref{eq:cs_cr_old}. With the same compression ratio, we have $M=S/N$. 

The advantageous of the proposed architecture is therefore evident. The actual elements of the sensing matrix are reduced by a factor of $N^2$, this alleviates the system complexity. Terahertz images are now reconstructed column by column in dimension $N\times 1$ rather than seek the sparse solution in the whole signal space $N^2$. However, more advanced algorithms are required to properly utilize the structures of THz images.

\section{The BMMV Modeled Sparse Bayesian Learning Algorithm}
\begin{figure}[!ht]
\centerline{\includegraphics[width=.3\linewidth]{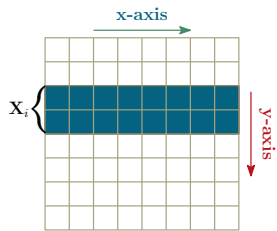}}
\caption{The block multiple measurement vector (BMMV) model}
\label{fig:bsbl_thz}
\end{figure}

In the scan-based CS-THz systems, we could recover the image column-by-column or take the columns as a whole and recover the image using multiple columns simultaneously. The latter is also called the multiple measurement vector (MMV) model. In practice, we observed that the adjacent columns of the THz image are highly correlated and each column may also exhibit a block sparse structure, as shown in Fig. \ref{fig:bsbl_thz}.
Exploiting such correlation information could bring additional performance improvements\cite{Zhang2012a}. 

The proposed algorithm was motivated by \cite{Zhang2013} and it was based on the Block Sparse Bayesian Learning (BSBL) framework\cite{Zhang2012a}. We also used the Fast Marginalized Likelihood Maximization (FMLM) method\cite{Tipping2003,Liu2012} to find efficient update formulas for the parameters in the Bayesian model.

\subsection{The Empirical Bayesian Model}
The signal $\ve{X}$ is divided into $g$ blocks with the $i$th ($i\in\{1,\ldots,g\}$) block denoted by $\ve{X}_i\in\mathbb{C}^{d_i\times N}$, where $d_i$ is the block size in the y-axis. The prior of the $i$th block $\ve{X}_i$ is modeled via,
\begin{equation}\label{eq:xi}
p(\ve{X}_i; \gamma_i) = \mathcal{MN}(\ve{x}_i;\ve{0},\gamma_i\ve{I}_{d_i},\ve{I}_N), 
\end{equation}
where
\begin{equation}
	\mathcal{MN}(\ve{X}_i;\bm{\mu}_i,\ve{U}_i,\ve{V}_i) = \frac{\exp\left(-\frac{1}{2}\mathrm{Tr}\left[\ve{V}_i^{-1}(\ve{X}_i - \bm{\mu}_i)^T\ve{U}_i^{-1}(\ve{X}_i - \bm{\mu}_i)\right]\right)}{(2\pi)^{d_iN/2}\abs{\ve{V}_i}^{d_i/2}\abs{U_i}^{N/2}} \nonumber
\end{equation}
is the matrix variate distribution with mean $\bm{\mu}_i$. The parameter $\ve{U}_i$ and $\ve{V}_i$ model the covariance of the row (y-axis) and column (x-axis) vectors of the signal block $\ve{X}_i$. In \eqref{eq:xi}, we modeled the variance of the block $\ve{X}_i$ by a single parameter $\gamma_i$. It can be viewed as the averaged variance of all the signals in the $i$th block, denoted by,
\begin{equation}
	\gamma_i \triangleq \frac{1}{d_iN} \mathrm{Tr}[\ve{U}_i\otimes \ve{V}_i].
\end{equation}
The learning process of $\gamma_i$ automatically determines the relevance of the $i$th block\cite{Tipping2003}.
We further assume that inter blocks are not correlated, the prior of the signal $\ve{X}$ reads,
\begin{equation}\label{eq:x_model}
p(\ve{X}|\{\gamma_i\}) = \mathcal{MN}(\ve{X};\ve{0},\bm{\Gamma},\ve{I}_N),
\end{equation}
where $\bm{\Gamma}$ is a block diagonal matrix with the $i$th principle diagonal given by $\gamma_i\ve{I}_{d_i}$. 

The measurements $\ve{Y}$ are modeled as,
\begin{equation}\label{eq:y_model}
p(\ve{Y};\ve{X},\beta) = \mathcal{MN}(\ve{Y};\bm{\Phi}\ve{X},\beta^{-1}\ve{I}_M,\ve{I}_N),
\end{equation}
where we assume that the measurements noise are i.i.d. Gaussian with the precision parameter given by $\beta$.

Given the signal model \eqref{eq:x_model} and the measurements model \eqref{eq:y_model}, we can derive the posterior $p(\ve{X}|\ve{Y}; \{\gamma_i\}, \beta)$ and the likelihood $p(\ve{Y}|\{\gamma_i\}, \beta)$ using the Gaussian Identity\cite[Appendix D]{Mahler2007},
\begin{align}
p(\ve{X}|\ve{Y};\{\gamma_i\},\beta) &= \mathcal{MN}(\ve{X};\bm{\mu},\bm{\Sigma},\ve{I}_N), \\
p(\ve{Y}|\{\gamma_i\},\beta) &= \mathcal{MN}(\ve{Y};\ve{0},\ve{C},\ve{I}_N),
\end{align}
where
\begin{align}
\bm{\mu} &= \beta\bm{\Sigma}\bm{\Phi}^H\ve{Y}, \\
\bm{\Sigma} &= (\bm{\Gamma}^{-1} + \beta\bm{\Phi}^H\bm{\Phi})^{-1}, \\
	\ve{C} &= \beta^{-1}\ve{I}_M + \bm{\Phi}\bm{\Gamma}\bm{\Phi}^H. \label{eq:c0}
\end{align}
To estimate the parameters $\{\gamma_i\},\beta$, the Type II Maximum Likelihood method\cite{Tipping2001} was used, which leads to the following cost function,
\begin{align}
\mathcal{L}(\{\gamma_i\},\beta) &\triangleq -2\log p(\ve{Y};\{\gamma_i\},\beta), \\
 &= N\log \abs{\ve{C}} + \mathrm{Tr}\left[ \ve{Y}^H\ve{C}^{-1}\ve{Y} \right]. \label{eq:loglikelihood}
\end{align}


\subsection{Fast Marginalized Implementation}

We denoted by $\bm{\Phi}_i\in\mathbb{C}^{N\times d_i}$ the $i$th column block in $\bm{\Phi}$. Therefore, $\ve{C}$ in \eqref{eq:c0} can be rewritten as:
\begin{align}
\ve{C} &= \beta^{-1}\ve{I} + \sum_{m\neq i}
\bm{\Phi}_m\gamma_m\bm{\Phi}_m^H+
\bm{\Phi}_i\gamma_i\bm{\Phi}_i^H, \\
 &= \ve{C}_{-i} + \bm{\Phi}_i\gamma_i\bm{\Phi}_i^H, \label{eq:c}
\end{align}
where $\ve{C}_{-i} \triangleq \beta^{-1}\ve{I} + \sum_{m\neq i}
\bm{\Phi}_m\gamma_m\bm{\Phi}_m^H$.
Using the Woodbury Identity, \eqref{eq:loglikelihood} can be rewritten as:
\begin{align}
\mathcal{L} =& N\log\abs{\ve{C}_{-i}} + \mathrm{Tr}\left[\ve{Y}^H\ve{C}_{-i}^{-1}\ve{Y}\right] 
\nonumber \\
 &+ N\log\abs{\ve{I}_{d_i} + \gamma_i\ve{s}_i} - \mathrm{Tr}\left[\ve{q}_i^H(\gamma_i^{-1}\ve{I}_{d_i} + \ve{s}_i)^{-1}\ve{q}_i\right], 
\nonumber \\
 =& \mathcal{L}(-i) + \mathcal{L}(i), \nonumber
\end{align}
where 
$\ve{s}_i\triangleq \bm{\Phi}_i^H\ve{C}_{-i}^{-1}\bm{\Phi}_i$, 
$\ve{q}_i\triangleq\bm{\Phi}_i^H\ve{C}_{-i}^{-1}\ve{Y}$ and 
\begin{align}
\mathcal{L}(i) &\triangleq N\log\abs{\ve{I}_{d_i} + \gamma_i\ve{s}_i} -
\mathrm{Tr}\left[ \ve{q}_i^H(\gamma_i^{-1}\ve{I}_{d_i} + \ve{s}_i)^{-1}\ve{q}_i\right]
\end{align}
which only depends on $\gamma_i$. By optimizing $\mathcal{L}(i)$, we could find the greedy update rule for $\gamma_i$,
\begin{equation}\label{eq:gamma_0}
\gamma_i =
\frac{1}{d_iN}\mathrm{Tr}\left[\ve{s}_i^{-1}(\ve{q}_i\ve{q}_i^H - \ve{s}_i)\ve{s}_i^{-1}\right].
\end{equation}

The proposed algorithm (denoted as BSBL-FM-MMV) is given in Fig. \ref{algo:bsbl-fm}. 
\begin{figure}[h!]
\centering
\begin{algorithmic}[1]
	\algrule
\Procedure{BSBL-FM-MMV}{$\ve{Y}$,$\bm{\Phi}$,$\eta$}
\State Outputs: $\ve{X},\bm{\Sigma},\bm{\gamma}$
\State Initialize $\beta^{-1}= 0.01\norm{\ve{Y}}_\mathcal{F}^2$
\State Initialize $\{\ve{s}_i\}$, $\{\ve{q}_i\}$
\While{not converged}
\State Calculate $\tilde{\gamma_i} = \frac{1}{d_iN}\mathrm{Tr}\left[\ve{s}_i^{-1}(\ve{q}_i\ve{q}_i^H - \ve{s}_i)\ve{s}_i^{-1}\right]$
\State Calculate the $\Delta \mathcal{L}(i) = \mathcal{L}(\tilde{\gamma_i}) - \mathcal{L}(\gamma_i), \forall i$
\State Update the $\hat{i}$th block s.t. $\Delta\mathcal{L}(\hat{i})=\min\{\Delta\mathcal{L}(i)\}_i$
\State Re-calculate $\bm{\mu},\bm{\Sigma},\{\ve{s}_i\},\{\ve{q}_i\}$
\State Re-calculate the convergence criterion
\EndWhile
\EndProcedure
	\algrule
\end{algorithmic}
\caption{The Proposed BSBL-FM-MMV Algorithm.}
\label{algo:bsbl-fm}
\end{figure}
Within each iteration, it only updates the block signal that attributes to the deepest descent of $\mathcal{L}(i)$. The detailed procedures on re-calculation of $\bm{\mu},\bm{\Sigma},\{\ve{s}_i\},\{\ve{q}_i\}$ were similar to \cite{Tipping2003}. The algorithm terminates when the  maximum normalized change of the cost function is small than a threshold $\eta$.

{\flushleft\bf Remark I:}
The proposed algorithm is a natural multiple measurement vector (MMV) extension to the BSBL-FM algorithm\cite{Liu2012}. The cost function of BSBL-FM\cite{Liu2012} is
\begin{equation}
\mathcal{L}'(\{\gamma_i\},\beta) = \log \abs{\ve{C}} + \ve{y}^T\ve{C}^{-1}\ve{y}, \label{eq:loglikelihood2}
\end{equation}
which is equivalent to \eqref{eq:loglikelihood} with the number of measurement vectors set to $1$. The proposed algorithm can also work in the complex-valued scenarios\footnote{This extension can also be derived by imposing the Circulant Symmetric properties\cite{adali2011complex} on the prior of the complex-valued signal $\{\ve{X}_i\}$ and the measurements $\ve{Y}$.} by simply replacing all the transpose operator $(\cdot)^T$ to the conjugate-transpose operator $(\cdot)^H$ as in \cite{Wipf2007}. The proposed algorithm could iteratively add relevant blocks into the model. It is most efficient when the signal is truly block-sparse across multiple measurements.

{\flushleft\bf Remark II:}
$\beta$ can be estimated from \eqref{eq:loglikelihood}. In practice, people often treat it as a regularizer\cite{Liu2012} and assign some specific values to it. Similar to \cite{Liu2012}, We select $\beta^{-1}=0.01\norm{\ve{Y}}_\mathcal{F}^2$ in the experiments.

{\flushleft\bf Remark III:}
The proposed algorithm can recover the signal in the transformed domain, i.e.,
\begin{equation}
\ve{Y} = (\bm{\Phi}\ve{F}) \ve{A},
\end{equation}
where $\ve{F}$ may be a Discrete Fourier Transform (DFT) Matrix and $\ve{A}\triangleq\{\bm{\alpha}_1,\ldots,\bm{\alpha}_N\}$ are the Fourier coefficients. Then we can recover the original signal $\hat{\ve{X}}$ by $\hat{\ve{X}} = \ve{F}\hat{\ve{A}}$.

\section{Experiments}

\subsection{The Experiments Setup}
We compared the proposed algorithm with SPG-L1\cite{Berg2008}, SL0\cite{Mohimani2008}, SPG-Group\cite{Berg2008}, BZAP\cite{liu2012efficient} and SPG-MMV\cite{Berg2008}. SPG-L1 and SL0 are typical $\ell_1$ and smoothed $\ell_0$ minimization solvers respectively. SPG-Group and BZAP are block/group based CS algorithms. SPG-MMV is a multiple-measurement-vector (MMV) extension to SPG-L1\cite{Berg2008}. 
The SPG-L1, SL0, SPG-Group and BZAP recovered the signal column-by-column, i.e., $\ve{y}_i=(\bm{\Phi}\ve{F})\ve{a}_i$ and $\hat{\ve{x}_i} = \ve{F}\hat{\ve{a}_i}$. All the algorithms were tuned to their optimal performances in the experiments.

Two performance indexes are used in the experiments. One is the Signal-to-Noise Ratio (SNR), defined as 
\begin{equation}
\text{SNR} \triangleq 10\log_{10} \frac{\norm{\ve{X}}_\mathcal{F}^2}{\norm{\hat{\ve{X}}-\ve{X}}_\mathcal{F}^2},
\end{equation}
where $\ve{\hat{X}}$ is the recovered signal of the true signal $\ve{X}$. The second index is the average CPU time. The computer used in the experiments has a 2.9GHz CPU and 16G RAM.
The experiments runs over $50$ trials and within each trial the sensing matrix was regenerated and the performance indexes were calculated.

\subsection{Validation using Synthetic THz Data}
\begin{figure}[!htb]
	\centering
	\subfloat[THz Image S0]{\includegraphics[width=.32\linewidth]{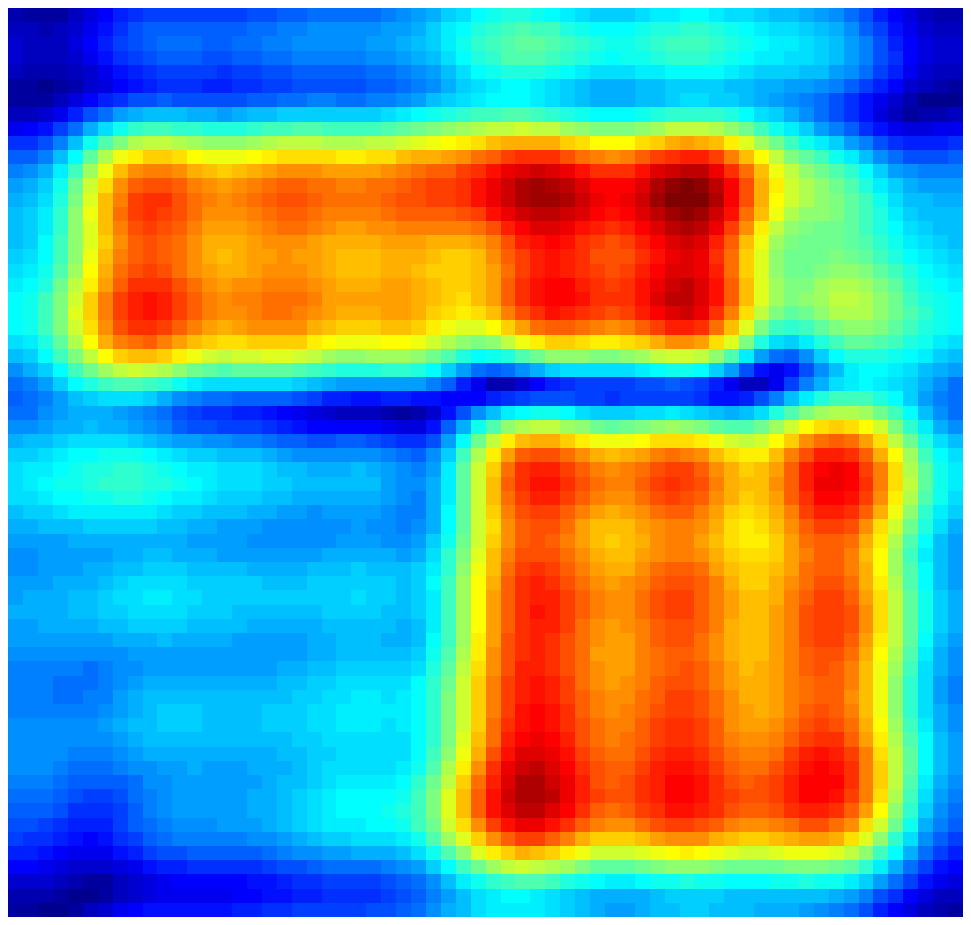}}
	\subfloat[THz Image S1]{\includegraphics[width=.32\linewidth]{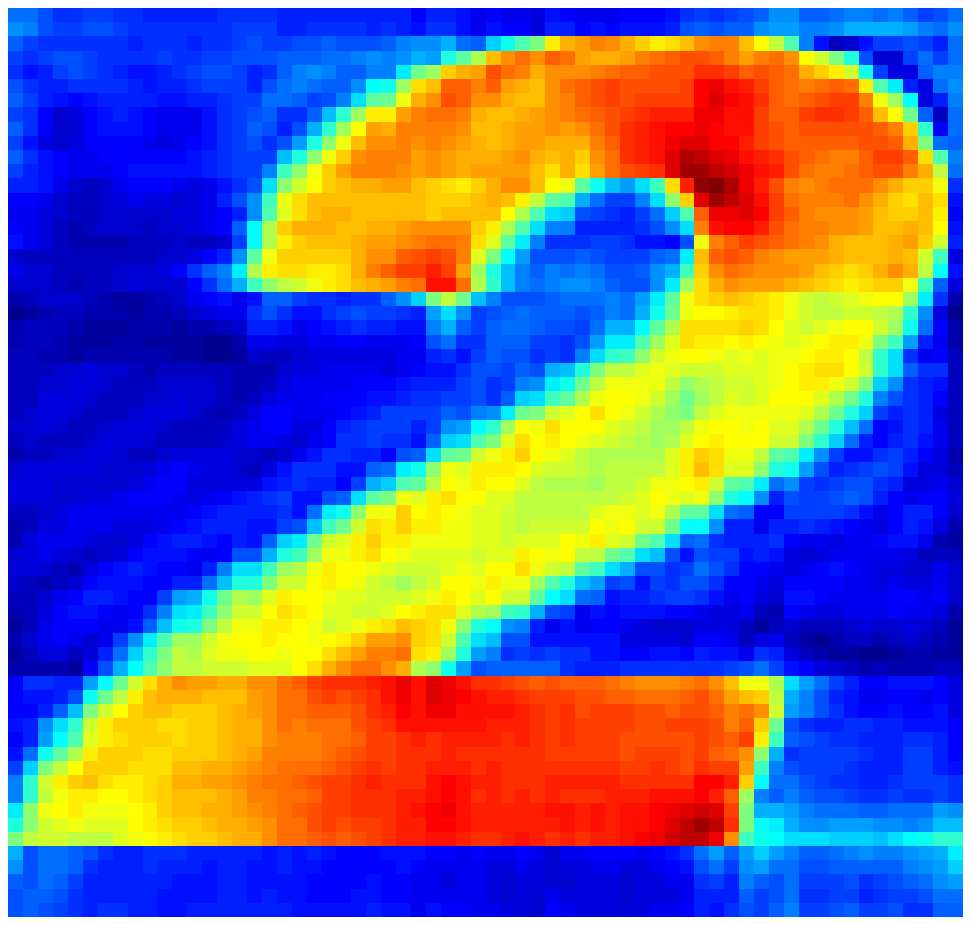}}
	\subfloat[THz Image S2]{\includegraphics[width=.32\linewidth]{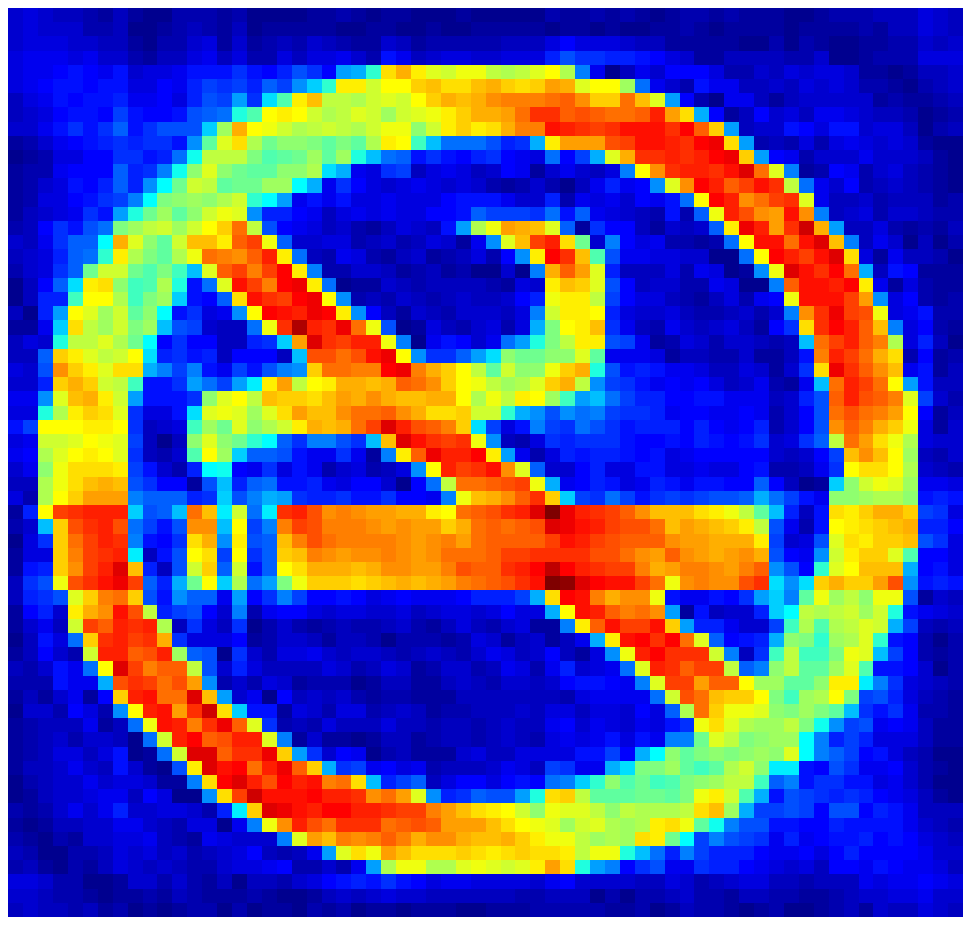}}
	\caption{The Synthetic THz images. The images are denoted by (from left to right) s0, s1 and s2 respectively.}
	\label{fig:thz_synth}
\end{figure}

The synthetic THz data are generated from low-pass filtered basic geometry shapes, as shown in Fig. \ref{fig:thz_synth}.
The THz images are denoted by S0 (a), S1 (b) and S2 (c) and the size of each image is fixed to $N=64$.
We evaluate the recovery performances by varying the compression ratios and adopting different forms of the sensing matrices.

\subsubsection{Performances with Varying Compression Ratios}
In this experiment, the sensing matrix was fixed to the complex-valued Gaussian matrices where the real and the image part of each entry $\phi_{ij}$ were i.i.d sampled from a standard Gaussian distribution. The compression ratios were varied from $\mathrm{CR}=0.5$ to $\mathrm{CR}=0.9$.

\begin{figure}[!htb]
	\centering
	\subfloat[THz Image S0]{\includegraphics[width=.32\linewidth]{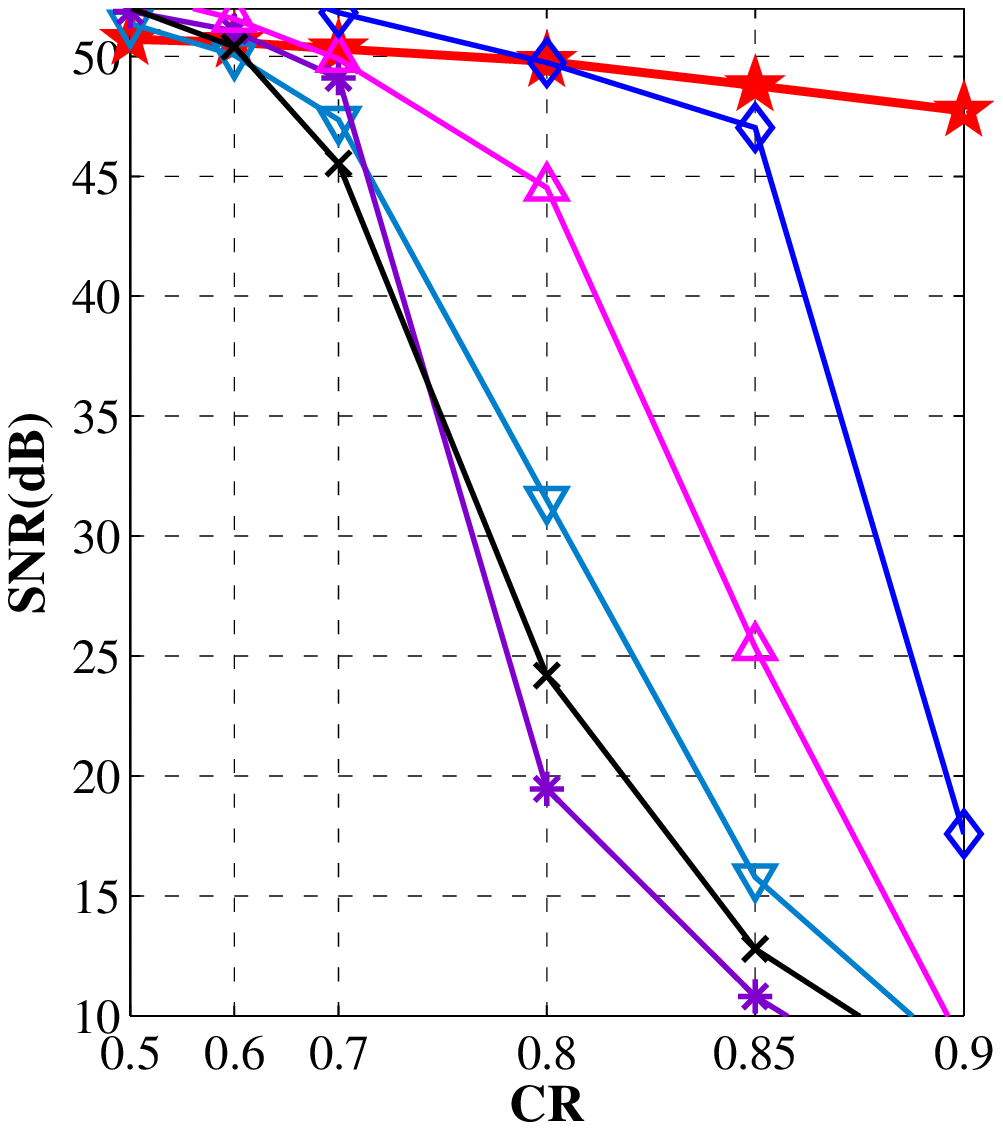}}
	\subfloat[THz Image S1]{\includegraphics[width=.32\linewidth]{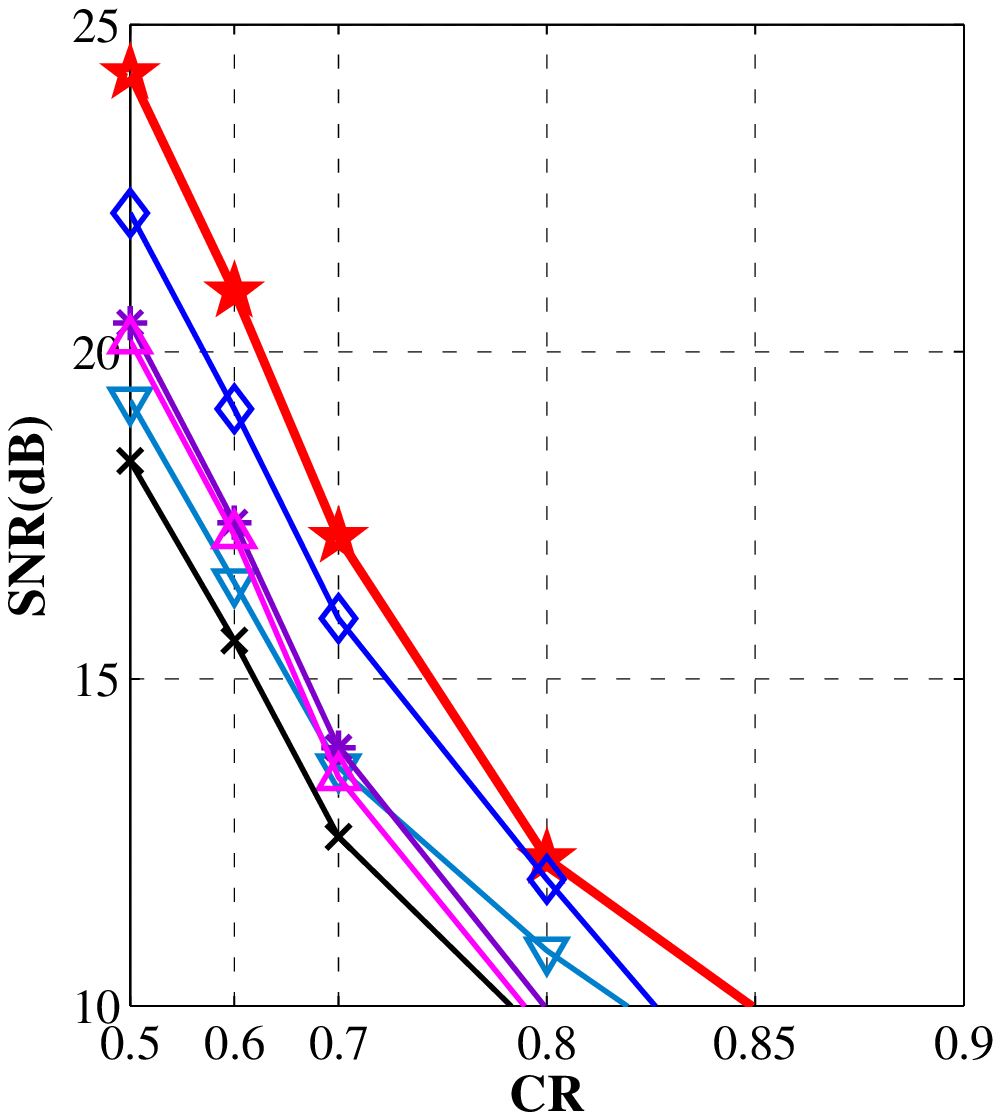}}
	\subfloat[THz Image S2]{\includegraphics[width=.32\linewidth]{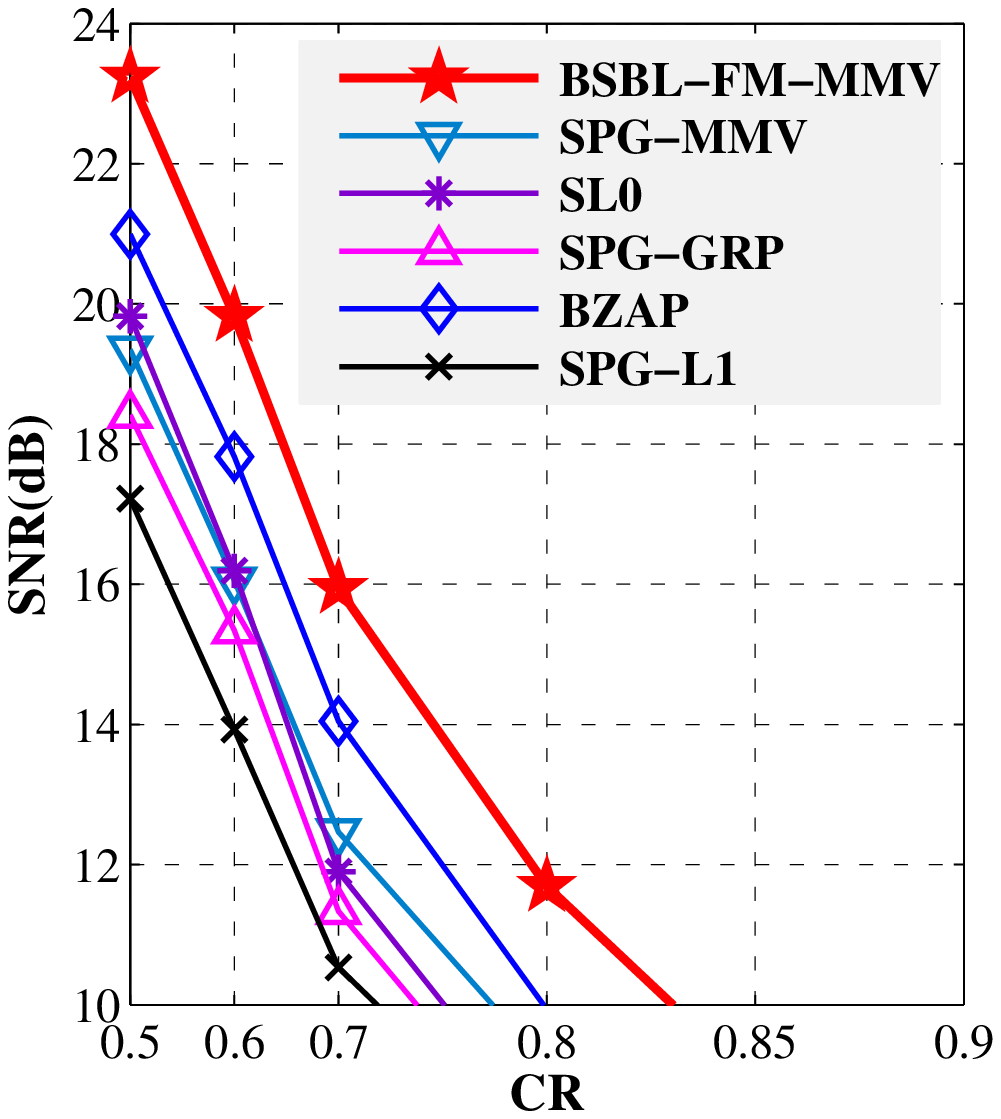}}
	\caption{The recovery performances on the synthetic dataset with varying compression ratios.}
	\label{fig:cr}
\end{figure}

From Fig. \ref{fig:cr}, we found that the proposed algorithm, BSBL-FM-MMV, showed the best performance in recovering the synthetic THz dataset. It is the only algorithm that fidelity recovered the S0 THz image with compression ratio $\mathrm{CR}=0.9$. 

\begin{figure}[!htb]
	\centering
	\subfloat[CPU Speed-up]{\includegraphics[width=.48\linewidth]{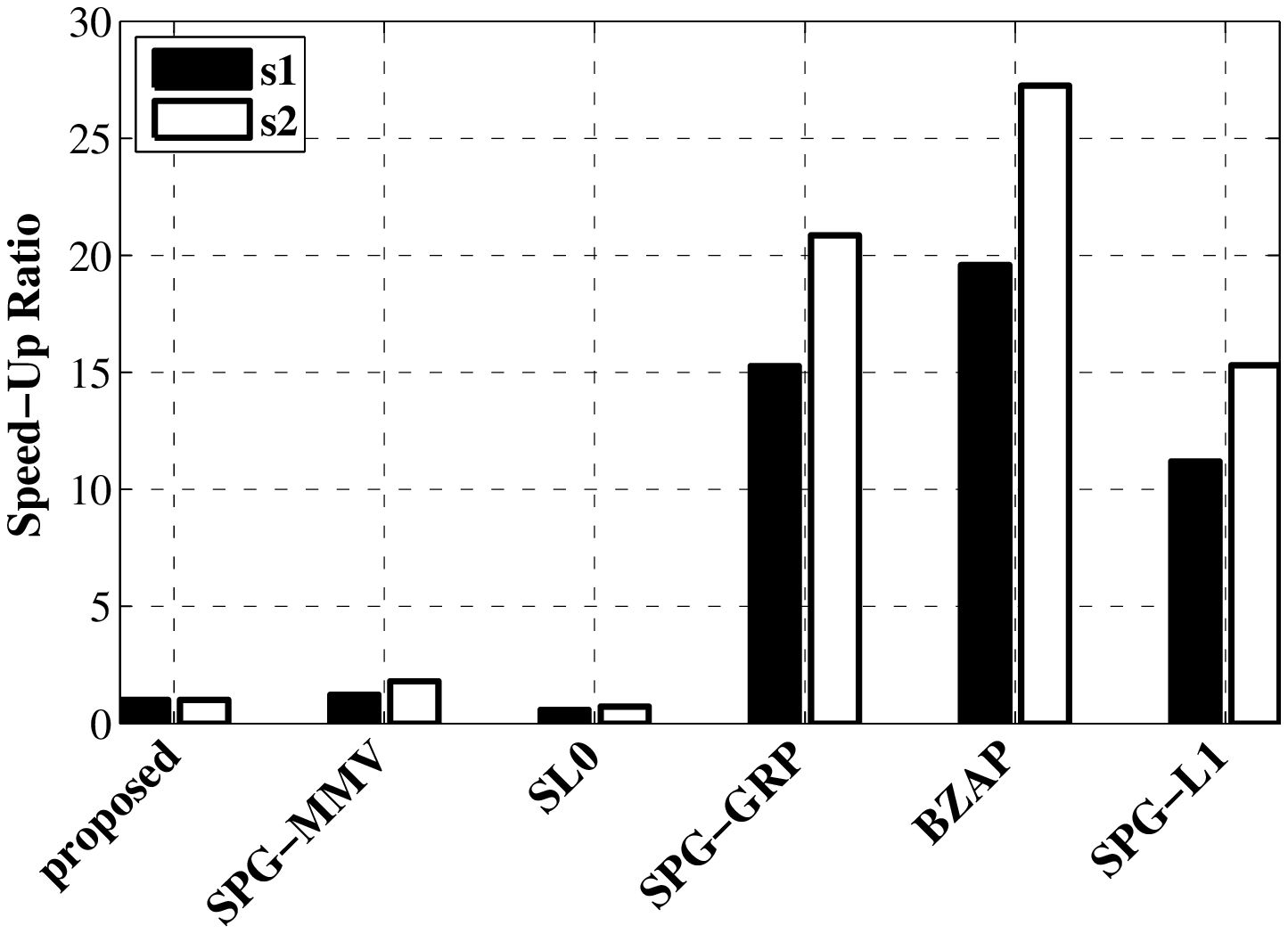}}
	\subfloat[SNR Improvement]{\includegraphics[width=.48\linewidth]{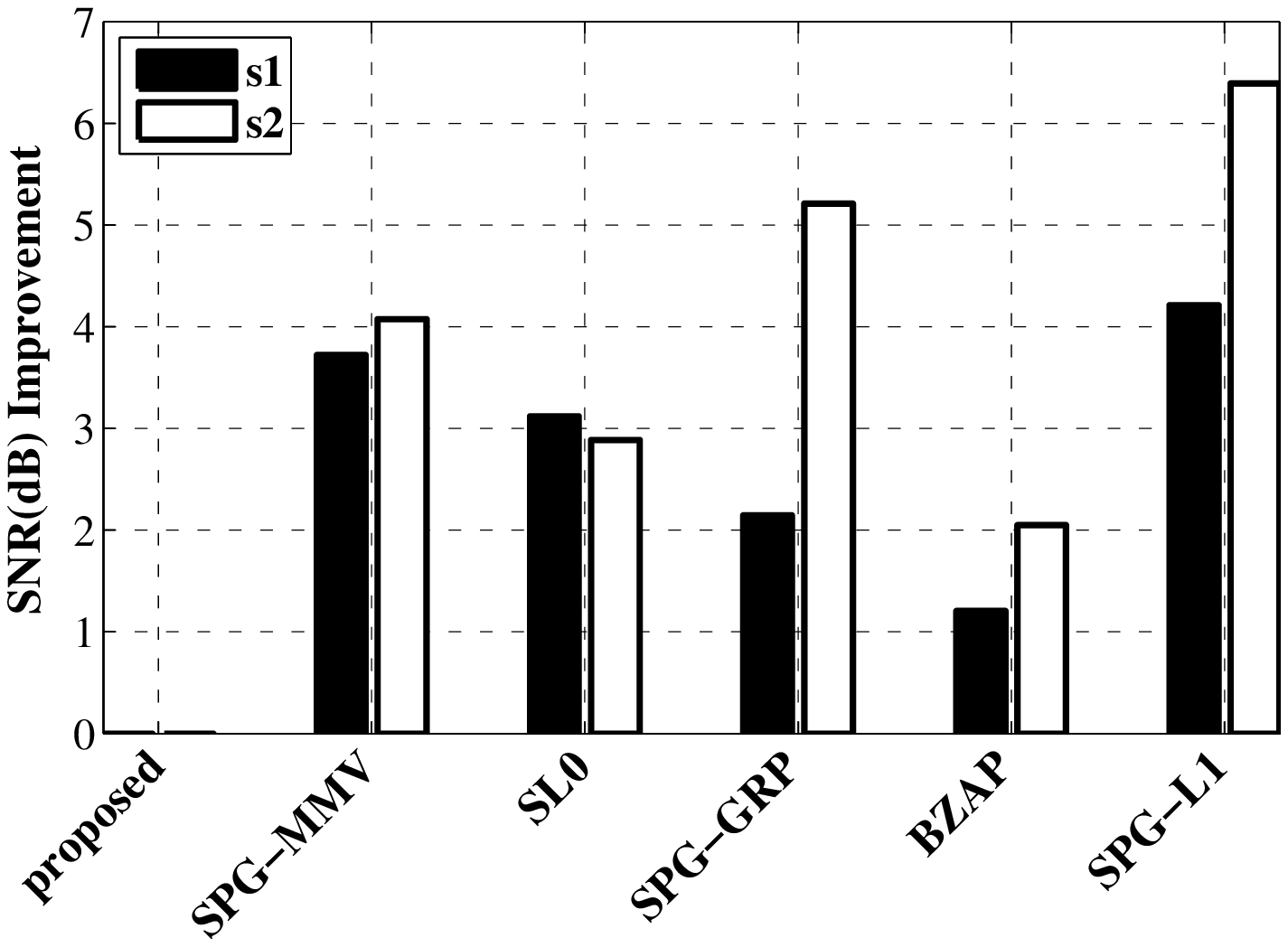}}
	\caption{The CPU speed-up and the SNR improvements in utilizing the BSBL-FM-MMV algorithm}
	\label{fig:speedup}
\end{figure}

The CPU speed-up ratios, defined as the normalized average CPU time with respect to the proposed algorithm, and the SNR improvements of utilizing the BSBL-FM-MMV algorithm were shown in Fig. \ref{fig:speedup}. In this experiment we fixed $\mathrm{CR}=0.8$.
The proposed algorithm was both superior in recovery performance and efficiency. It was slower than SL0 but its SNR was $3$dB better. The BSBL-FM-MMV algorithm was $15$ times faster than SPG-Group and $20$ times faster than BZAP while still yielded improved SNR performances.

\subsubsection{Performances with Different Forms of the Sensing Matrices}
\begin{figure}[!htb]
	\centering
	\subfloat[THz Image S1]{\includegraphics[width=.48\linewidth]{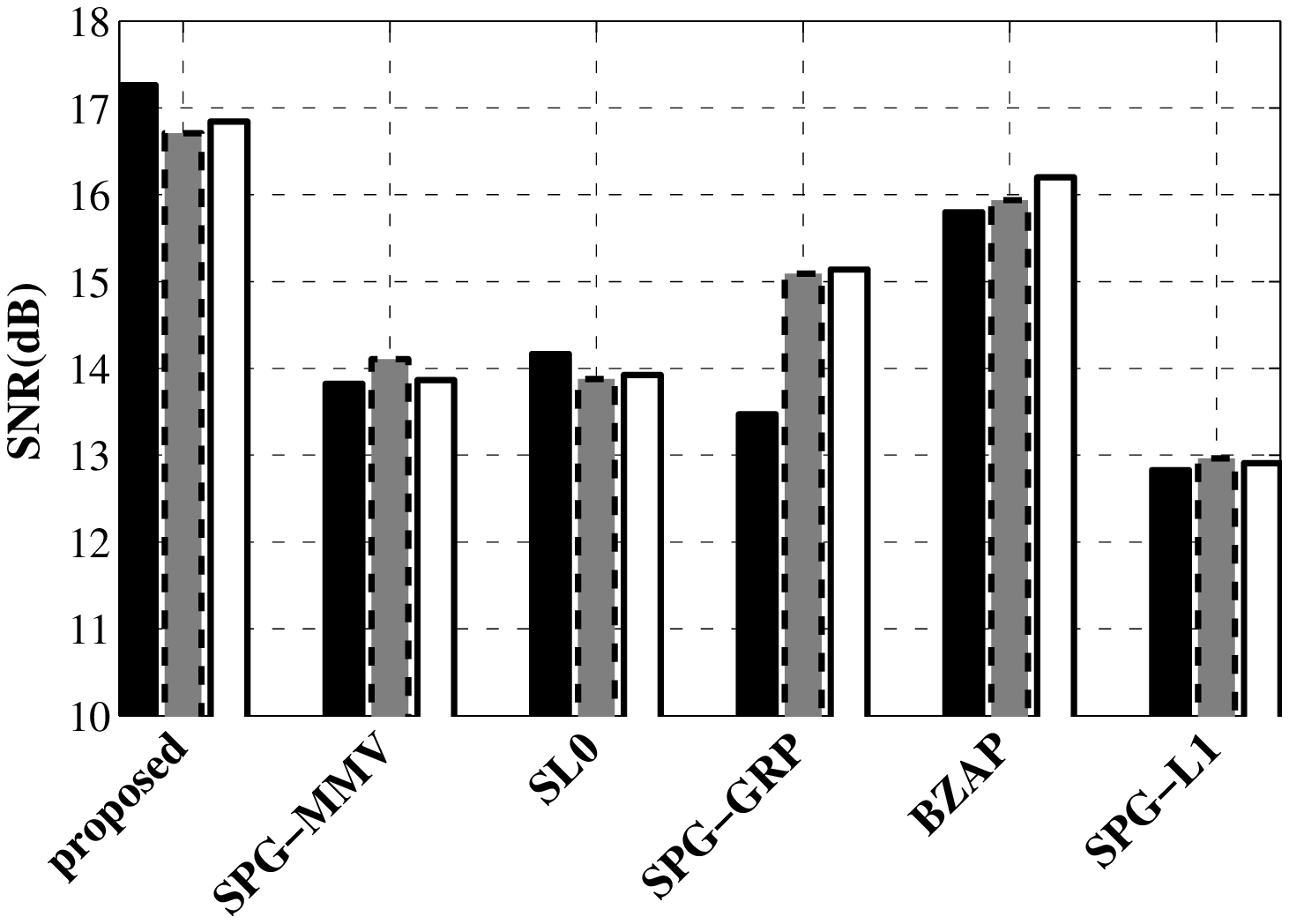}}
	\subfloat[THz Image S2]{\includegraphics[width=.48\linewidth]{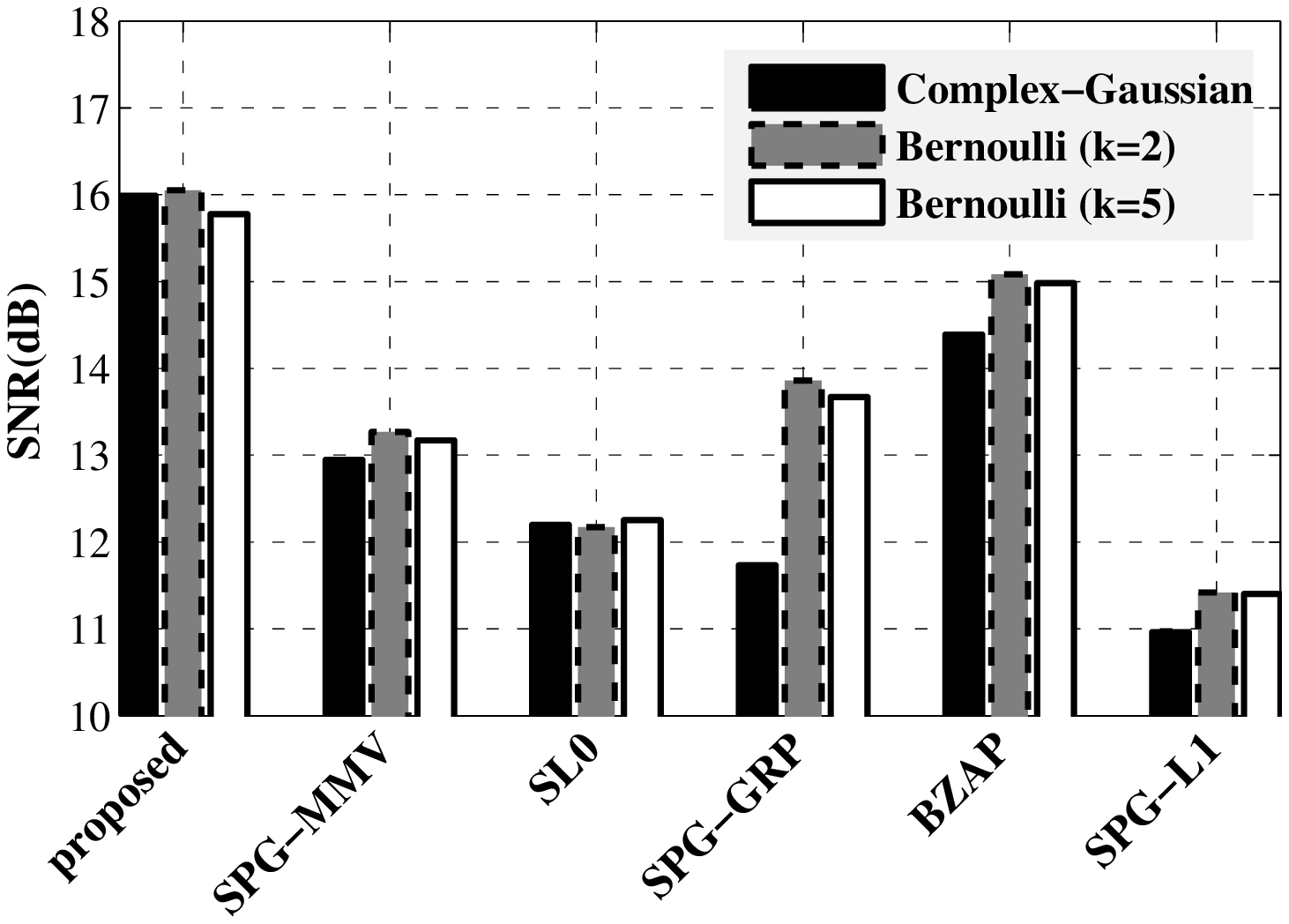}}
	\caption{The reconstruction performances using different sensing matrices.}
	\label{fig:phi}
\end{figure}

We analyzed the performances of CS recovery by utilizing different sensing matrices. Besides the complex-valued Gaussian matrix used in the previous section, we also adopt the Bernoulli sensing matrix whose entries consisted of either $0$ or $1$. The number of the non-zero entries each column was fixed to $k$ where $k=2$ and $k=5$ were used in this experiment. We fixed the $\mathrm{CR}=0.7$ and the results were shown in Fig. \ref{fig:phi}.
The SNR property of the proposed algorithm was not affected by the choice of the sensing matrix. In practical applications, the scan based CS-THz system can utilize the metamaterial to build the complex-valued sensing matrix. 

\subsection{Validation using Real-Life Terahertz Image}

\subsubsection{An illustrative Example}
\begin{figure}[htbp]
\centering
\includegraphics[width=.6\linewidth]{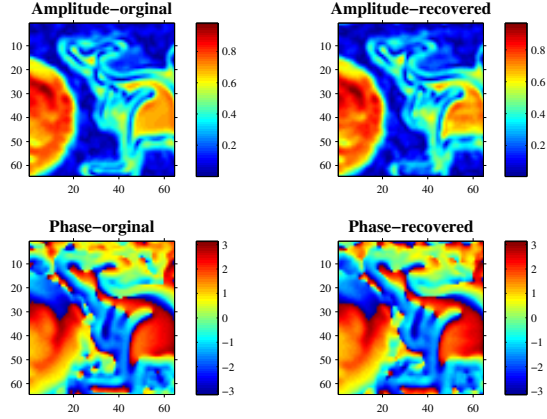}
\caption{Experiment on real-life THz image with $N=64$, $\mathrm{CR}=0.5$. 
}%
\label{fig:thz_01}
\end{figure}

The real-life THz data used in this section was the same as in \cite{Yu2012,Popescu2010}. 
In this experiment, we resized the image to $N=64$ and fixed $\mathrm{CR}=0.5$. The sensing matrix $\bm{\Phi}$ was a randomly generated Bernoulli matrix with exactly $k=5$ non-zero entries each column. A reconstruction of the complex-valued Terahertz image using the proposed algorithm was given in Fig. \ref{fig:thz_01}.
The proposed algorithm recovered the original image in only $1.06$s with $\text{SNR}=21.08$dB. The CS algorithm reported in \cite{Yu2012} applied on the same Terahertz image takes $20$min on an Intel Core 2 Duo Processor (3GHz), achieved maximum $11$dB SNR with the same experiment settings ($N=64$, $\mathrm{CR}=0.5$). Both the reconstruction time and the image quality of the proposed algorithm were superior to the results in \cite{Yu2012}.

\subsubsection{Performances with Varying $N$}
\begin{figure}[!htb]
	\centering
	\subfloat[]{\includegraphics[width=.48\linewidth]{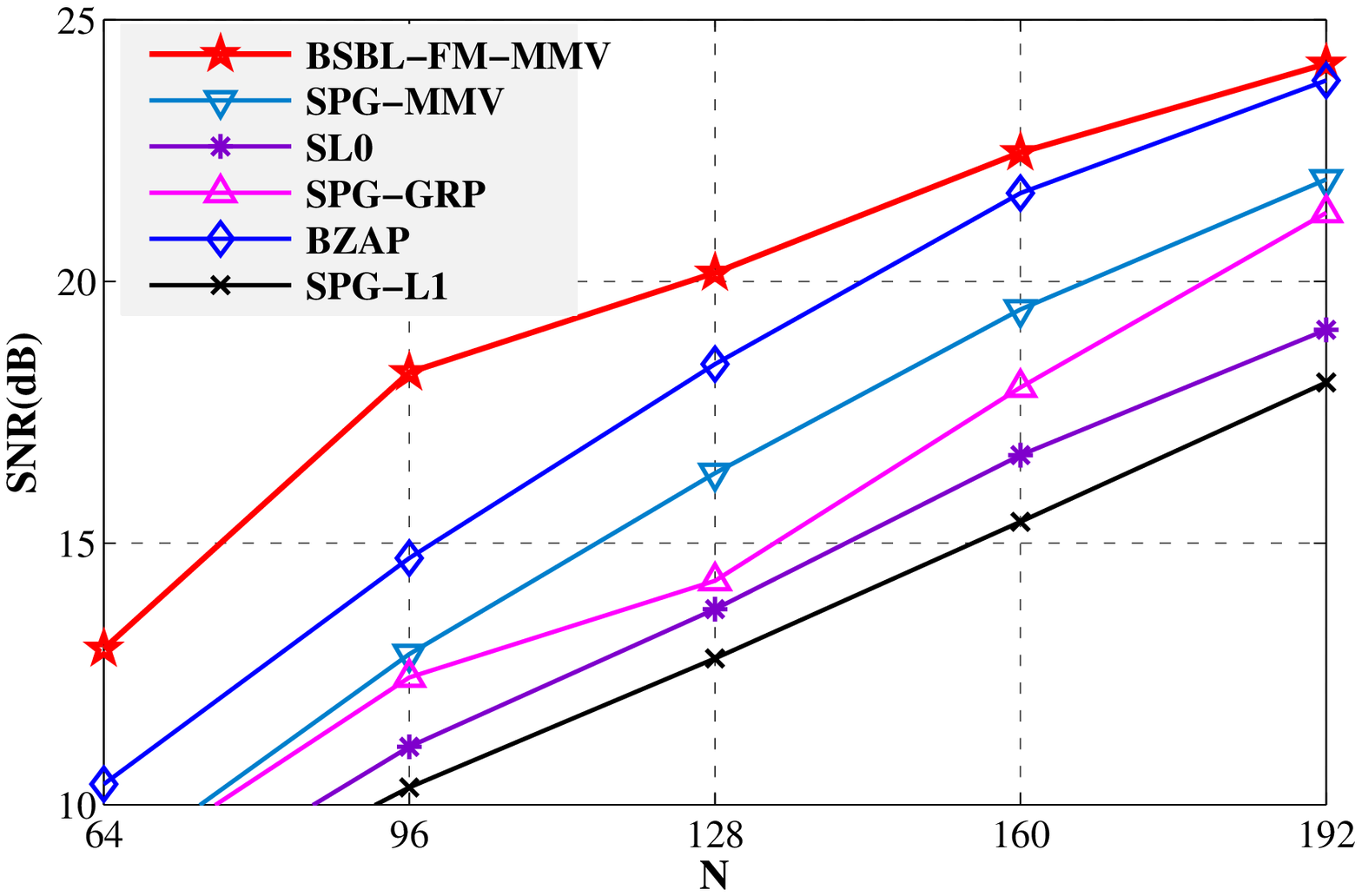}}
	\subfloat[]{\includegraphics[width=.48\linewidth]{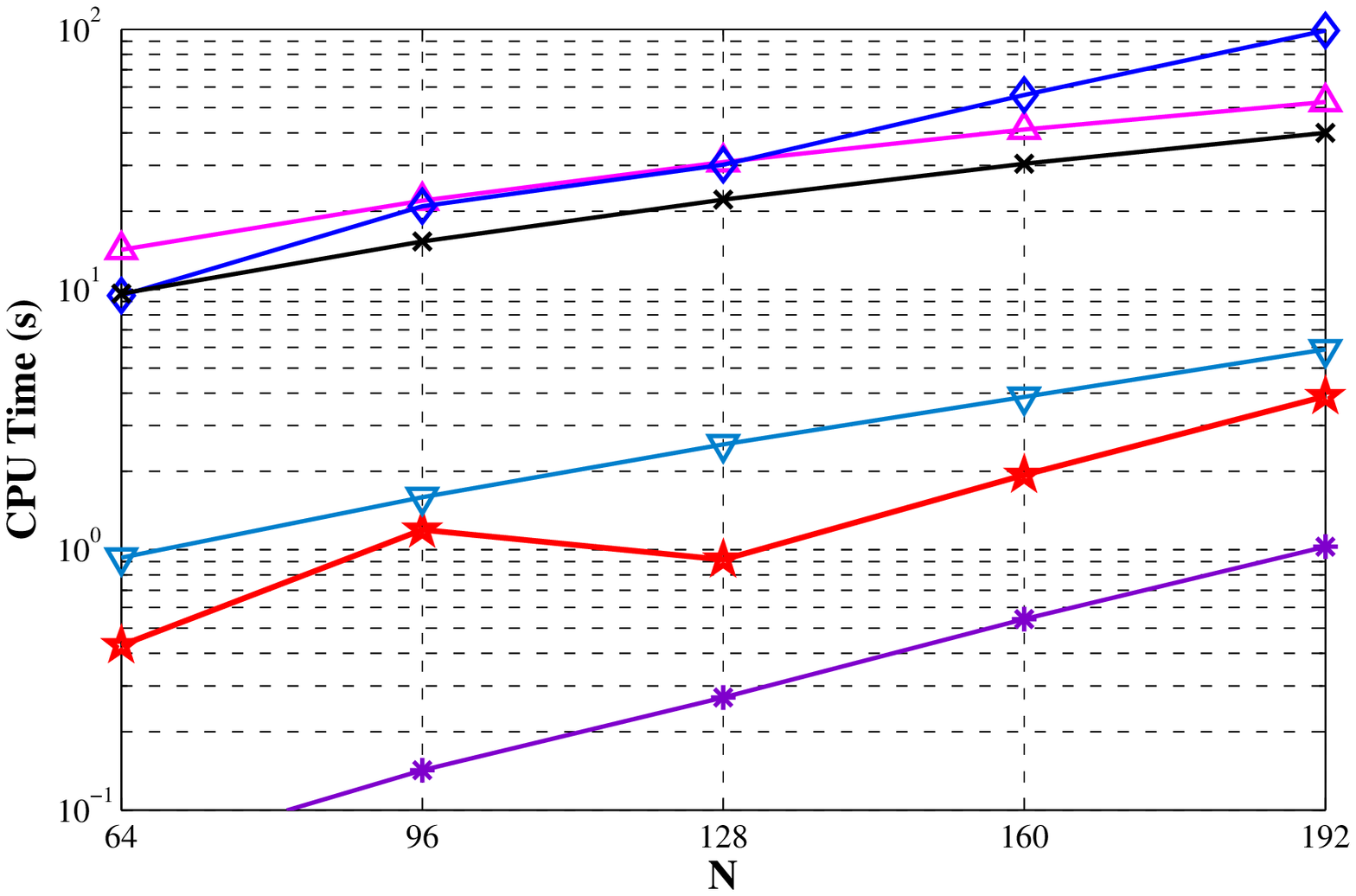}}
	\caption{The reconstruction performance with varying $N$.}%
	\label{fig:thz_02}
\end{figure}

In this experiment, we fixed $\mathrm{CR}=0.7$ and assessed the reconstruction performance with varying $N$, the results were shown in Fig. \ref{fig:thz_02}.
The proposed algorithm showed the best SNR in this simulation. The proposed algorithm also scaled well with different image sizes. We should also note that the block size $b_i$ can be tuned for practical applications to achieve better reconstruction results and efficiency.

\subsubsection{Performances with Varying Compression Ratio $\mathrm{CR}$}
\begin{figure}[htbp]
\centering
\includegraphics[width=.5\linewidth]{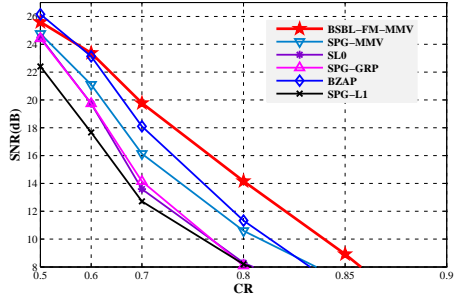}
\caption{The reconstruction performance with varying $\mathrm{CR}$.}%
\label{fig:thz_03}
\end{figure}

In this experiment, we fixed $N=128$ and assessed the reconstruction performance with varying $\mathrm{CR}$. The results were shown in Fig. \ref{fig:thz_03}.

These algorithms showed similar results when $\mathrm{CR}=0.5$. However, the proposed algorithm was more superior especially with larger compression ratios. Combining together the scan based CS-THz architecture and the BSBL-FM-MMV algorithm, we could fidelity recover a $128\times128$ sized THz image with compression ratio $\mathrm{CR}=0.7$. The average $\text{SNR}$ was $20$dB and the CPU time was only $2.08$s. 
\section{Conclusion}
We presented a novel CS-THz imaging architecture that compressed sensing the THz image column by column. The sensing matrix can be reduced by a factor of $N^2$. Therefore, it alleviated the system complexity. In order to utilize the structures of the THz image, an advanced CS recovery algorithm was proposed. It showed improved performance compared with those traditional CS solvers. Combining the scan based CS-THz system and the proposed CS recovery algorithm, we could efficiently and fidelity sense the object using THz waveforms. These properties are very attractive for practical Terahertz imaging systems.

\section*{Acknowledgment}
This work was supported in part by the National Natural Science Foundation of China under Grant 61101186.



\bibliographystyle{IEEEtran}
\bibliography{bsbl}

\begin{thebibliography}{10}
\providecommand{\url}[1]{#1}
\csname url@samestyle\endcsname
\providecommand{\newblock}{\relax}
\providecommand{\bibinfo}[2]{#2}
\providecommand{\BIBentrySTDinterwordspacing}{\spaceskip=0pt\relax}
\providecommand{\BIBentryALTinterwordstretchfactor}{4}
\providecommand{\BIBentryALTinterwordspacing}{\spaceskip=\fontdimen2\font plus
\BIBentryALTinterwordstretchfactor\fontdimen3\font minus
  \fontdimen4\font\relax}
\providecommand{\BIBforeignlanguage}[2]{{%
\expandafter\ifx\csname l@#1\endcsname\relax
\typeout{** WARNING: IEEEtran.bst: No hyphenation pattern has been}%
\typeout{** loaded for the language `#1'. Using the pattern for}%
\typeout{** the default language instead.}%
\else
\language=\csname l@#1\endcsname
\fi
#2}}
\providecommand{\BIBdecl}{\relax}
\BIBdecl

\bibitem{Shen2012}
H.~Shen, ``Compressed sensing on terahertz imaging,'' Ph.D. dissertation,
  University of Liverpool, 2012.

\bibitem{Duarte2008}
M.~F. Duarte, M.~A. Davenport, D.~Takhar, J.~N. Laska, T.~Sun, K.~F. Kelly, and
  R.~G. Baraniuk, ``Single-pixel imaging via compressive sampling,'' \emph{IEEE
  Signal Processing Magazine}, vol. March, pp. 83--91, 2008.

\bibitem{Berg2008}
E.~V.~D. Berg and M.~P. Friedlander, ``Probing the pareto frontier for basis
  pursuit solutions,'' \emph{SIAM Journal on Scientific Computing}, vol. 31(2),
  pp. 890--912, 2008.

\bibitem{Candes2006}
E.~Candes, J.~Romberg, and T.~Tao, ``Robust uncertainty principles: exact
  signal reconstruction from highly incomplete frequency information,''
  \emph{Information Theory, IEEE Transactions on}, vol.~52, no.~2, pp. 489 --
  509, feb. 2006.

\bibitem{Candes2008a}
E.~Candes and M.~Wakin, ``An introduction to compressive sampling,''
  \emph{Signal Processing Magazine, IEEE}, vol.~25, no.~2, pp. 21 --30, march
  2008.

\bibitem{li2013compressive}
S.~Li, X.~Zhou, B.~Ren, H.-J. Sun, and X.~Lv, ``A compressive sensing approach
  for synthetic aperture imaging radiometers,'' \emph{Progress In
  Electromagnetics Research}, vol. 135, pp. 583--599, 2013.

\bibitem{tonouchi2007cutting}
M.~Tonouchi, ``Cutting-edge terahertz technology,'' \emph{Nature photonics},
  vol.~1, no.~2, pp. 97--105, 2007.

\bibitem{ohrstrom2010technical}
L.~{\"O}hrstr{\"o}m, A.~Bitzer, M.~Walther, and F.~J. R{\"u}hli, ``Technical
  note: terahertz imaging of ancient mummies and bone,'' \emph{American journal
  of physical anthropology}, vol. 142, no.~3, pp. 497--500, 2010.

\bibitem{hunt2013metamaterial}
J.~Hunt, T.~Driscoll, A.~Mrozack, G.~Lipworth, M.~Reynolds, D.~Brady, and D.~R.
  Smith, ``Metamaterial apertures for computational imaging,'' \emph{Science},
  vol. 339, no. 6117, pp. 310--313, 2013.

\bibitem{kuznetsov2012matrix}
S.~A. Kuznetsov, A.~G. Paulish, A.~V. Gelfand, P.~A. Lazorskiy, and V.~N.
  Fedorinin, ``Matrix structure of metamaterial absorbers for multispectral
  terahertz imaging,'' \emph{Progress In Electromagnetics Research}, vol. 122,
  pp. 93--103, 2012.

\bibitem{Reinhard2013}
B.~Reinhard, O.~Paul, and M.~Rahm, ``Metamaterial-based photonic devices for
  terahertz technology,'' \emph{Selected Topics in Quantum Electronics, IEEE
  Journal of}, vol.~19, no.~1, pp. 8\,500\,912--8\,500\,912, 2013.

\bibitem{Zhang2013}
Z.~Zhang, T.-P. Jung, S.~Makeig, B.~D. Rao, and Z.~Pi, ``Spatiotemporal sparse
  bayesian learning with applications to compressed sensing of multichannel eeg
  for wireless telemonitoring and brain-computer interfaces,''
  \emph{(submitted) IEEE Trans. on Neural Systems and Rehabilitation
  Engineering}, 2013.

\bibitem{Zhang2012a}
Z.~Zhang and B.~Rao, ``Extension of {SBL} algorithms for the recovery of block
  sparse signals with intra-block correlation,'' \emph{Signal Processing, IEEE
  Transactions on}, vol.~61, no.~8, pp. 2009--2015, 2013.

\bibitem{Tipping2003}
M.~E. Tipping and A.~C. Faul, ``Fast marginal likelihood maximisation for
  sparse bayesian models,'' in \emph{Proceedings of the Ninth International
  Workshop on Artificial Intelligence and Statistics}, C.~M. Bishop and B.~J.
  Frey, Eds., Key West, FL, 2003, pp. 3--6.

\bibitem{Liu2012}
B.~Liu, Z.~Zhang, and H.~Fan, ``Fast marginalized block sparse bayesian
  learning algorithm,'' \emph{Arxiv:1211.4909}, 2012.

\bibitem{Mahler2007}
R.~P.~S. Mahler, \emph{Statistical Multisource-Multitarget Information
  Fusion}.\hskip 1em plus 0.5em minus 0.4em\relax Artech House, 2007.

\bibitem{Tipping2001}
M.~E. Tipping, ``Sparse bayesian learning and the relevance vector machine,''
  \emph{Journal of Machine Learning Research}, vol.~1, pp. 211--244, 2001.

\bibitem{adali2011complex}
T.~Adali, P.~Schreier, and L.~Scharf, ``Complex-valued signal processing: the
  proper way to deal with impropriety,'' \emph{Signal Processing, IEEE
  Transactions on}, vol.~59, no.~11, pp. 5101--5125, 2011.

\bibitem{Wipf2007}
D.~P. Wipf and B.~Rao, ``An empirical bayesian strategy for solving the
  simultaneous sparse approximation problem,'' \emph{Signal Processing, IEEE
  Transactions on}, vol.~55, no.~7, pp. 3704 --3716, july 2007.

\bibitem{Mohimani2008}
G.~H. Mohimani, M.~Babaie-Zadeh, and C.~Jutten, ``A fast approach for
  overcomplete sparse decomposition based on smoothed $\ell^0$ norm,''
  \emph{IEEE Transactions on Signal Processing}, vol. 57 (1), pp. 1--13, 2008.

\bibitem{liu2012efficient}
J.~Liu, J.~Jin, and Y.~Gu, ``Efficient recovery of block sparse signals via
  zero-point attracting projection,'' in \emph{Acoustics, Speech and Signal
  Processing (ICASSP), 2012 IEEE International Conference on}.\hskip 1em plus
  0.5em minus 0.4em\relax IEEE, 2012, pp. 3333--3336.

\bibitem{Yu2012}
S.~Yu, A.~S. Khwaja, and J.~Ma, ``Compressed sensing of complex-valued data,''
  \emph{Signal Processing}, vol.~92, pp. 357--362, 2012.

\bibitem{Popescu2010}
D.~Popescu and A.~Hellicar, ``Point spread function estimation for a terahertz
  imaging system,'' \emph{EURASIP J. Adv. Signal Process}, 2010.

\end{thebibliography}

%
%
%
\end{document}